\documentclass[11pt,a4paper]{article}
\usepackage[hyperref]{acl2020}
\usepackage{times}
\usepackage{tikz}
\usepackage{subfig}
\usepackage{xcolor}
\usepackage{tcolorbox}
\usepackage{latexsym}

\usepackage{multirow}
\usepackage{booktabs}
\usepackage{pgfplots}
\usepackage{amssymb}
\usepackage{xspace}

\usetikzlibrary{plotmarks}
\usetikzlibrary{backgrounds}
\usetikzlibrary{fit}
\usetikzlibrary{decorations} 
\usetikzlibrary{decorations.markings}
\usetikzlibrary{positioning}

\definecolor{myblue}{RGB}{31,120,180}
\definecolor{mygreen}{RGB}{51,160,44}
\definecolor{myred}{RGB}{227,26,28}

\definecolor{ugreen}{rgb}{0,0.5,0}
\definecolor{iyellow}{RGB}{255,250,205}
\definecolor{ipurple}{RGB}{230,230,250}

\usepackage{microtype}

\aclfinalcopy 

\title{Does Multi-Encoder Help? A Case Study on Context-Aware \\ Neural Machine Translation}

\author{
  Bei Li$^1$,
  Hui Liu$^1$,
  Ziyang Wang$^1$,
  Yufan Jiang$^1$, \\
  \textbf{Tong Xiao$^{1,2}$\thanks{\xspace\xspace Corresponding author.}},
  \textbf{Jingbo Zhu$^{1,2}$},
  \textbf{Tongran Liu$^3$},
  \textbf{Changliang Li$^4$} \\
  $^{1}$NLP Lab, Northeastern University, Shenyang, China\\
  $^{2}$NiuTrans Research, Shenyang, China \\
  $^{3}$CAS Key Laboratory of Behavioral Science, Institute of Psychology, CAS, Beijing, China \\
  $^{4}$Kingsoft AI Lab, Beijing, China \\
  {\tt
        \{libei\_neu,jiangyufan2018\}@outlook.com, 
  }\\
  {\tt
        \{huiliu,wangziyang\}@stumail.neu.edu.cn,
  }\\
  {\tt
        \{xiaotong,zhujingbo\}@mail.neu.edu.com,
  } \\
  {\tt
        liutr@psych.ac.cn,lichangliang@kingsoft.com
  } \\
}

\date{}

\begin{document}
\maketitle
\begin{abstract}
In encoder-decoder neural models, multiple encoders are in general used to represent the contextual information in addition to the individual sentence. In this paper, we investigate multi-encoder approaches in document-level neural machine translation (NMT). Surprisingly, we find that the context encoder does not only encode the surrounding sentences but also behaves as a noise generator. This makes us rethink the real benefits of multi-encoder in context-aware translation - some of the improvements come from robust training. We compare several methods that introduce noise and/or well-tuned dropout setup into the training of these encoders. Experimental results show that noisy training plays an important role in multi-encoder-based NMT, especially when the training data is small. Also, we establish a new state-of-the-art on IWSLT Fr-En task by careful use of noise generation and dropout methods.
\end{abstract}

\section{Introduction}

Sentence-level neural machine translation (NMT) systems ignore the discourse phenomena and encode the individual source sentences with no use of contexts. In recent years, the context-aware models which learn contextual information from surrounding sentences have shown promising results in generating consistent and coherent translations \cite{zhang-etal-2018-improving,voita-etal-2018-context,kim-etal-2019-document,voita-etal-2019-good,bawden-etal-2018-evaluating,miculicich-etal-2018-document,maruf-haffari-2018-document,maruf-etal-2019-selective}.

There are two common approaches to incorporating contexts into NMT: the simple way is to concatenate the context and the current sentence to form a context-aware input sequence \cite{Agrawal2018ContextualHI,tiedemann-scherrer-2017-neural}, whereas a more widely-used approach utilizes additional neural networks to encode context sentences \cite{journals/corr/JeanLFC17,voita-etal-2018-context,zhang-etal-2018-improving}. Here we name the former as the single-encoder approach and name the latter as the multi-encoder approach.
However, large-scale document corpora are not easily available. Most context-aware NMT systems are evaluated on small datasets and significant BLEU improvements are reported \cite{wang-etal-2017-exploiting-cross,zhang-etal-2018-improving,tu-etal-2018-learning}. In our experiments, we find that the improvement persists if we feed pseudo sentences into the context encoder, especially when we train the system on small-scale data. A natural question here is: \textit{How much does the improvement come from the leverage of contextual information in multi-encoder}?

\begin{figure*}[htb]
  \centering
  \tikzstyle{prenode} = [rounded corners=2pt,inner sep=4pt,minimum height=1.8em,minimum width=5em,draw,fill=blue!10!white]
  \tikzstyle{curnode} = [rounded corners=2pt,inner sep=4pt,minimum height=1.8em,minimum width=5em,draw,fill=red!30!white]
  \tikzstyle{denode} = [rounded corners=2pt,inner sep=4pt,minimum height=9em,minimum width=7em,draw,fill=black!5!white]
  \tikzstyle{denode1} = [rounded corners=2pt,inner sep=4pt,minimum height=5em,minimum width=18em,draw,fill=black!5!white]
  \tikzstyle{attnode} = [rounded corners=3pt,inner sep=4pt,minimum height=1.5em,minimum width=4.5em,draw,fill=green!30!white]
  \tikzstyle{standard} = [rounded corners=3pt,thick]
  \subfloat[Outside]
  {
    \centering
    \begin{tikzpicture}
        \node [prenode,anchor=west] (pre) at (0,0) {\footnotesize{$\mathrm{Encoder_{c}}$}};
        \node [curnode,anchor=west] (cur) at ([xshift=1.0em]pre.east) {\footnotesize{$\mathrm{Encoder_{s}}$}};
        \node [denode,anchor=south west] (decoder) at ([xshift=1.0em]cur.south east) {$\mathrm{}$};
        \node [attnode,anchor=south west] (att1) at ([xshift=-0.8em,yshift=6em]pre.south east) {\footnotesize{$\mathrm{Attention}$}};
        \node [attnode,anchor=south] (att2) at ([yshift=5em]decoder.south) {\footnotesize{$\mathrm{Attention}$}};

        \node [anchor=north] (null) at ([yshift=-2em]pre.south) {\footnotesize{}};

        \node [anchor=north] (labelc) at ([yshift=-0.8em]pre.south) {\footnotesize{Context}};
        \node [anchor=north] (labels) at ([yshift=-0.8em]cur.south) {\footnotesize{Source}};
        \node [anchor=north] (labelt) at ([xshift=2em,yshift=-0.8em]decoder.south) {\footnotesize{Target}};
        \node [anchor=south] (label1) at ([yshift=0.6em]pre.north) {\footnotesize{$H_{\mathrm{c}}$}};
        \node [anchor=south] (label2) at ([yshift=0.7em]cur.north) {\footnotesize{$H_{\mathrm{s}}$}};
        \node [anchor=south] (label3) at ([yshift=0.7em]att1.north) {\footnotesize{$H_{\mathrm{c^{'}}}$}};
        \node [anchor=south] (label4) at ([xshift=-1.3em,yshift=0.5em]decoder.south) {\footnotesize{$\mathrm{Decoder}$}};

        \node [anchor=west] (plus1) at ([xshift=0.7em,yshift=0em]label2.east) {$\mathbf{\oplus}$};
        \node [anchor=south] (dot1) at ([yshift=1.8em]att2.north) {$\mathbf{...}$};
        \node [anchor=south] (dot2) at ([xshift=2em,yshift=0.2em]decoder.south) {$\mathbf{...}$};

        \draw [->,thick] ([yshift=-0.1em]labelc.north) -- ([yshift=-0.2em]pre.south);
        \draw [->,thick] ([yshift=-0.1em]labels.north) -- ([yshift=-0.2em]cur.south);
        \draw [->,thick] ([yshift=-0.1em]labelt.north) -- ([xshift=2em,yshift=-0.2em]decoder.south);

        \draw [->,thick] ([yshift=0.1em]decoder.north) -- ([yshift=1em]decoder.north);
        \draw [->,thick] ([yshift=0.1em]pre.north) -- ([yshift=0.8em]pre.north);
        \draw [->,thick] ([yshift=0.1em]cur.north) -- ([yshift=0.8em]cur.north);
        \draw [->,thick] ([yshift=0.1em]att1.north) -- ([yshift=0.8em]att1.north);
        \draw [->,thick] ([yshift=0.1em]att2.north) -- ([yshift=2em]att2.north);

        \draw [->,thick] ([yshift=0.1em]label1.north) .. controls +(north:0.58) and +(south:0.78) ..([yshift=-0.2em]att1.south);
        \draw [->,thick] ([yshift=0.1em]label1.north) .. controls +(north:0.48) and +(south:0.73) ..([xshift=-2em,yshift=-0.2em]att1.south);
        \draw [->,thick] (label2.north) -- ([yshift=1.9em]label2.north);

        \draw [->,standard] (label2.east) -- ([xshift=1.0em]label2.east);
        \draw [->,standard] (label3.east) -- ([xshift=3.4em]label3.east) -- ([xshift=3.4em,yshift=-5.3em]label3.east);

        \draw [->,thick] ([xshift=-0.1em]plus1.east) .. controls +(north:0) and +(south:0.7) ..([yshift=-0.2em]att2.south);
        \draw [->,thick] ([xshift=-0.1em]plus1.east) .. controls +(north:0) and +(south:0.7) ..([xshift=-2em,yshift=-0.2em]att2.south);
        \draw [->,thick] ([xshift=2em,yshift=1em]decoder.south) -- ([xshift=2em,yshift=4.8em]decoder.south);

    \end{tikzpicture}
  }
  \hfill
  \subfloat[Inside]
  {
    \centering
    \begin{tikzpicture}
        \node [prenode,anchor=west] (pre) at (0,0) {\footnotesize{$\mathrm{Encoder_{c}}$}};
        \node [curnode,anchor=west] (cur) at ([xshift=8.0em]pre.east) {\footnotesize{$\mathrm{Encoder_{s}}$}};
        \node [denode1,anchor=south west] (decoder) at ([xshift=0em,yshift=4em]pre.south west) {$\mathrm{}$};
        \node [attnode,anchor=south] (att1) at ([xshift=-4em,yshift=1.5em]decoder.south) {\footnotesize{$\mathrm{Attention_{c}}$}};
        \node [attnode,anchor=south] (att2) at ([xshift=4em,yshift=1.5em]decoder.south) {\footnotesize{$\mathrm{Attention_{s}}$}};

        \node [anchor=north] (null) at ([yshift=-2em]pre.south) {\footnotesize{}};

        \node [anchor=south] (label1) at ([yshift=0.6em]pre.north) {\footnotesize{$H_{\mathrm{c}}$}};
        \node [anchor=south] (label2) at ([yshift=0.7em]cur.north) {\footnotesize{$H_{\mathrm{s}}$}};
        \node [anchor=south] (label4) at ([xshift=-7em,yshift=3.5em]decoder.south) {\footnotesize{$\mathrm{Decoder}$}};
        \node [anchor=center] (plus1) at ([xshift=4em,yshift=0.5em]att1.north) {$\mathbf{\oplus}$};
        \node [anchor=north] (labelc) at ([yshift=-0.8em]pre.south) {\footnotesize{Context}};
        \node [anchor=north] (labels) at ([yshift=-0.8em]cur.south) {\footnotesize{Source}};
        \node [anchor=north] (labelt) at ([yshift=-0.8em]decoder.south) {\footnotesize{Target}};
        \node [anchor=south] (dot1) at ([yshift=0.1em]plus1.north) {$\mathbf{...}$};
        \node [anchor=south] (dot2) at ([yshift=-0.2em]decoder.south) {$\mathbf{...}$};

        \draw [->,thick] ([yshift=-0.1em]labelc.north) -- ([yshift=-0.2em]pre.south);
        \draw [->,thick] ([yshift=-0.1em]labels.north) -- ([yshift=-0.2em]cur.south);
        \draw [->,thick] ([yshift=-0.1em]labelt.north) -- ([yshift=-0.2em]decoder.south);
        \draw [->,thick] ([yshift=0.1em]pre.north) -- ([yshift=0.8em]pre.north);
        \draw [->,thick] ([yshift=0.1em]cur.north) -- ([yshift=0.8em]cur.north);
        \draw [->,thick] ([yshift=0.3em]decoder.south) .. controls +(north:0.3) and +(south:0.4) ..([xshift=2em,yshift=-0.2em]att1.south);
        \draw [->,thick] ([yshift=0.3em]decoder.south) .. controls +(north:0.3) and +(south:0.4) ..([xshift=-2em,yshift=-0.2em]att2.south);

        \draw [->,thick] ([yshift=0em]label1.north) .. controls +(north:0.3) and +(south:0.5) ..([xshift=-2em,yshift=-0.2em]att1.south);
        \draw [->,thick] ([yshift=0em]label1.north) .. controls +(north:0.5) and +(south:0.5) ..([yshift=-0.2em]att1.south);
        \draw [->,thick] ([yshift=0em]label2.north) .. controls +(north:0.3) and +(south:0.5) ..([xshift=2em,yshift=-0.2em]att2.south);
        \draw [->,thick] ([yshift=0em]label2.north) .. controls +(north:0.5) and +(south:0.5) ..([yshift=-0.2em]att2.south);
        \draw [->,standard] ([yshift=0.1em]att1.north) -- ([yshift=0.5em]att1.north) -- ([xshift=3.5em,yshift=0.5em]att1.north);
        \draw [->,standard] ([yshift=0.1em]att2.north) -- ([yshift=0.5em]att2.north) -- ([xshift=-3.5em,yshift=0.5em]att2.north);
        \draw [->,thick] ([yshift=0.1em]decoder.north) -- ([yshift=1em]decoder.north);
        \draw [->,thick] ([yshift=-0.2em]plus1.north) -- ([yshift=0.3em]plus1.north);
    \end{tikzpicture}
  }

  \caption{An overview of two multi-encoder systems. In the Outside approach, $H_\mathrm{s}$ is the query and $H_\mathrm{c}$ is the key/value. In the Inside approach, $\mathrm{Target}$ is the query, $H_\mathrm{s}$ and $H_\mathrm{c}$ represent key/value.}\label{fig:architecture}
\end{figure*}
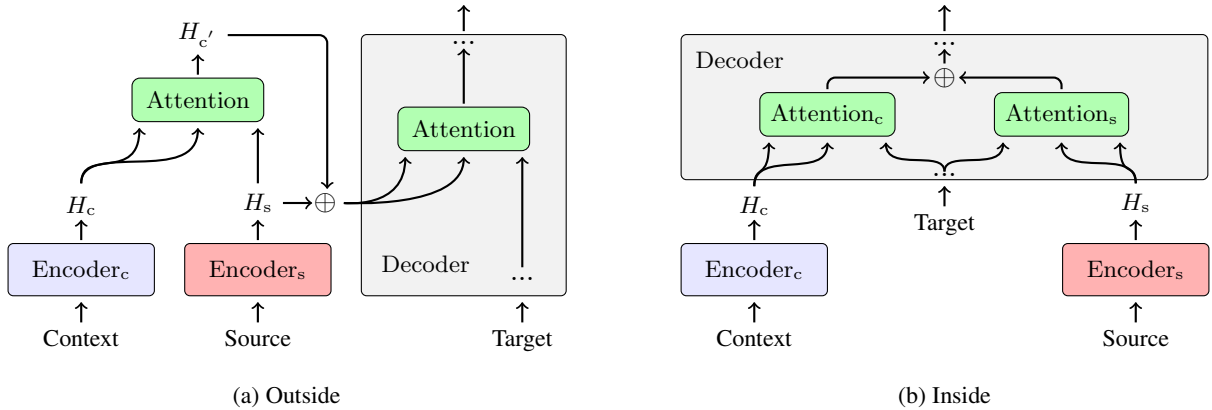

In this work, we aim to investigate what kinds of information that the context-aware model captures. We re-implement several widely used context-aware architectures based on the multi-encoder paradigm, and do an in-depth analysis to study whether the context encoder captures the contextual information.
By conducting extensive experiments on several document-level translation benchmarks, we observe that:
\begin{itemize}
  \item The BLEU gaps between sentence-level and context-aware models decrease when the sentence baselines are carefully tuned, e.g., proper use of dropout.
  \item The multi-encoder systems are insensitive to the context input. Even randomly sampled sentences can bring substantial improvements.
  \item The model trained with the correct context can achieve better performance during inference without the context input.
\end{itemize}

Our contribution is two folds: (\romannumeral1) We find that the benefit of the multi-encoder context-aware approach is not from the leverage of contextual information. Instead, the context encoder acts more like a noise generator to provide richer training signals. (\romannumeral2) The finding here inspires us to develop a simple yet effective training strategy: we add a Gaussian-noise to the encoder output, which can effectively alleviate the overfitting, especially on small datasets.

\section{Approaches to Incorporating Contexts into NMT}
\label{sec:system}

Here we describe two ways of introducing contextual information into NMT systems.

\subsection{The Single-Encoder Approach}

The input of the single-encoder system is the concatenation of the context sentences and the current sentence, with a special symbol inserted to distinguish them \cite{tiedemann-scherrer-2017-neural,Agrawal2018ContextualHI}. Then the extended sentence is fed into the standard Transformer. These systems may face the challenge of encoding extremely long inputs, resulting in inefficient computation.

\subsection{The Multi-Encoder Approach}

The multi-encoder models take the surrounding sentences as the context and employ an additional neural network to encode the context, that is, we have a source-sentence encoder and a context encoder. Figure \ref{fig:architecture} shows two methods of integrating the context into NMT in the multi-encoder paradigm. Next we show that most of the multi-encoder approaches \cite{voita-etal-2018-context,zhang-etal-2018-improving} are instances of the models described below.

\begin{itemize}
\item \textbf{Outside integration}. As shown in Figure \ref{fig:architecture}(a), the representations of the context and the current sentence are firstly transformed into a new representation by an attention network. Then the attention output and the source sentence representation are fused by a gated sum. 

\item \textbf{Inside integration}. Alternatively, the decoder can attend to two encoders respectively (Figure \ref{fig:architecture}(b)). Then, the gating mechanism inside the decoder is employed to obtain the fusion vector.
\end{itemize}

\section{Experimental Setup}

\subsection{Data and Settings}
We evaluated the document-level approaches on several publicly available datasets. For Chinese-English (Zh-En) and French-English (Fr-En), we used Ted talks from IWSLT15 and IWSLT16 \cite{cettoloEtAl:EAMT2012} evaluation campaigns as the training data. We validated on \textit{dev2010}, and tested on \textit{tst2010-2013} (Zh-En), \textit{tst2010} (Fr-En) respectively. For English-German (En-De), we evaluated on WMT18 task \footnote{We used the News-Commentary v14 as the train set}. For more convincing results, we also randomly sampled 500k/1M/2M/5M sentence pairs from the Chinese-English corpus provided by WMT\footnote{http://www.statmt.org/wmt19/translation-task.html} and test on \textit{newstest2017}. We preprocessed the sentences with Moses tokenizer\footnote{http://www.statmt.org/moses} except Chinese sentences and used byte pair encoding \cite{sennrich-subword-neural} with 32K merged operations to segment words into sub-word units. The Chinese sentences were word segmented by the tool provided
within NiuTrans \cite{xiao-etal-2012-niutrans}. For Fr-En and Zh-En tasks, we lowercased all sentences to obtain comparable results with previous work. We also conducted experiments on a larger English-Russian (En-Ru) dataset provided by \citet{voita-etal-2018-context}, consisting of 2M sentence pairs selected from publicly available OpenSubtitles2018 corpus. The data statistics of each language pair can be seen in Table \ref{tab:data}. We chose the Transformer-base model as the sentence-level baseline. The context encoder also used the same setting as the sentence-level baseline.

\begin{table}[t]
  \setlength{\tabcolsep}{3.0pt}
    \centering
    \begin{tabular}{c|r|r|r|r|r|r}
    \hline
    \multicolumn{1}{c|}{\multirow{2}{*}{\centering Lang.}} & \multicolumn{2}{c|}{Train} & \multicolumn{2}{c|}{Valid} & \multicolumn{2}{c}{Test} \\ \cline{2-7}
    \multicolumn{1}{c|}{} &
    \multicolumn{1}{c|}{doc.} & \multicolumn{1}{c|}{sent.} & \multicolumn{1}{c|}{doc.} & \multicolumn{1}{c|}{sent.} & \multicolumn{1}{c|}{doc.} & \multicolumn{1}{c}{sent.} \\
    \hline
    Zh-En & 1708 & 209K & 8   & 887  & 56  & 5473 \\
    \hline
    Fr-En & 1803 & 220K & 8   & 887  & 11  & 1664 \\
    \hline
    En-De & 8462 & 329K & 130 & 3004 & 122 & 2998 \\
    \hline
    En-Ru & - & 2M & - & 10k & - & 10k \\
    \hline
    \end{tabular}
    \caption{\label{tab:data} Details of datasets on different language pairs. }
    \label{tab:data}
  \end{table}

We used Adam \cite{kingma2014adam} for optimization, and trained the systems on a single TiTan V GPU\footnote{For En-Ru and Zh-En we trained models on 4 GPUs}.
The learning rate strategy was the same as that used in \citet{vaswani2017attention}. Our implementation was based on Fairseq \cite{ott-etal-2019-fairseq}. More details can be found in our repository\footnote {The source code is available at \url{https://github.com/libeineu/Context-Aware}}.

\section{Results and Discussion}

To study whether the context-encoder network captures contextual information in training, we present three types of context as the input of the context-encoder:
\begin{itemize}
  \item \textit{Context}: the previous sentence of the current sentence.
  \item \textit{Random}: a sentence consisting of words randomly sampled from the source vocabulary.
  \item \textit{Fixed}: a fixed sentence input for context-encoder.
\end{itemize}

\subsection{Baseline Selection}
Weight sharing \cite{voita-etal-2018-context} and two-stage training \cite{zhang-etal-2018-improving} strategies have been proven essential to build strong context-aware systems. The former shared the first N-1 blocks of context encoder with the source encoder, and the latter first trained a standard sentence-level Transformer
and finetuned the document-level Transformer with an extra context-encoder.
We first evaluated the importance of two training strategies for multi-encoder systems. We selected the multi-encoder with Outside integration (see Section \ref{sec:system}) as the context-aware model and trained systems with two training strategies on the En-De task respectively. As shown in Table \ref{tab:ablation}, we find that both two strategies outperform the sentence-level baseline by a large margin. The model with two-stage training performs slightly better than the weight-sharing system in terms of BLEU. To our surprise, the context-encoder with a single-layer can compete with a six-layers model. We suspect that this is because the training data is limited and we do not need a sophisticated model to fit it. Therefore, we choose the two-stage training and single-layer context-encoder for all experiments in the remainder of this paper.

\begin{table}[t]
  \small
  \begin{center}
  \begin{tabular}{c|ccccc}
  \toprule
  \textbf{System}&\textbf{Layers} & \textbf{WS} & \textbf{TS} & \textbf{BLEU} \\
  \midrule
  Sentence-level & -         & -          & -          & 28.9 \\
  \cmidrule(r){1-1} \cmidrule(l){2-5}
  \multirow{4}{*}{Outside \textit{Context}}
  & 6         & $\times$ & $\times$     & 28.5 \\
  & 6         & \checkmark & $\times$   & 29.3 \\
  & 6         & $\times$   & \checkmark & 29.6 \\
  & 1         & $\times$   & \checkmark & 29.4 \\

  \bottomrule
  \end{tabular}
  \end{center}
\caption{Comparison of context-aware model with two training strategies on En-De task. WS represents weight-sharing and TS represents two-stage training.}\label{tab:ablation}
\end{table}

\subsection{Results}

\begin{table*}[t!]
  \begin{center}
    \small
  \begin{tabular}{cccccccccc}
  \toprule
  \multicolumn{2}{c}{\multirow{2}{*}{\textbf{System}}}
  & \multicolumn{2}{c}{\textbf{Zh-En}}
  & \multicolumn{2}{c}{\textbf{Fr-En}}
  & \multicolumn{2}{c}{\textbf{En-De}}
  & \multicolumn{2}{c}{\textbf{En-Ru}} \\
  \cmidrule(lr){3-4} \cmidrule(lr){5-6} \cmidrule(lr){7-8} \cmidrule(lr){9-10}
  & &$p=0.1$ & $p=0.3$ & $p=0.1$ & $p=0.3$ & $p=0.1$ & $p=0.3$ & $p=0.1$ & $p=0.3$\\
  \midrule
  \multicolumn{2}{c|}{Sentence-level}
  & 18.0 & 19.7 & 36.5 & 36.9 & 28.9 & 30.2 & 30.3 & 31.1\\
  \multicolumn{2}{c|}{Single-encoder}
  & 18.1 & 19.1 & 36.2 & 37.3 & 28.5 & 30.2 & 30.4 & 31.2  \\
  \cmidrule(lr){1-2} \cmidrule(lr){3-4} \cmidrule(lr){5-6} \cmidrule(lr){7-8} \cmidrule(lr){9-10}
  \multirow{3}{*}{Inside}
   & \multicolumn{1}{|l|}{\textit{Context}}
   & 19.4 & 20.0 & 36.8 & 37.5 & 29.7 & 31.0 & 30.8 & 31.3  \\
   & \multicolumn{1}{|l|}{\textit{Random}}
   & 19.5 & 20.3 & 37.0 & 37.4 & 29.9 & 30.7 & 30.8 & 31.4 \\
   & \multicolumn{1}{|l|}{\textit{Fixed}}
   & 19.5 & 20.3 & 37.0 & 37.2 & 29.3 & 30.8 & 30.8 & 31.4 \\
  \cmidrule(lr){1-2} \cmidrule(lr){3-4} \cmidrule(lr){5-6} \cmidrule(lr){7-8} \cmidrule(lr){9-10}
  \multirow{3}{*}{Outside}
  & \multicolumn{1}{|l|}{\textit{Context}}
  & 19.4 & 19.8 & 36.8 & 37.4 & 29.4 & 30.7 & 30.9 & 31.1\\
  & \multicolumn{1}{|l|}{\textit{Random}}
  & 19.4 & 20.1 & 36.8 & 37.3 & 29.6 & 31.1 & 30.7 & 31.1\\
  & \multicolumn{1}{|l|}{\textit{Fixed}}
  & 19.4 & 20.0 & 36.7 & 37.2 & 29.5 & 31.1 & 30.8 & 31.1\\
  \bottomrule
  \end{tabular}
  \end{center}
\caption{The BLEU scores $[\%]$ of different context-aware models with three context inputs. We use $dropout=0.1$ and $dropout=0.3$ respectively.}\label{tab:main-results}
\end{table*}

Table \ref{tab:main-results} shows the results of several context-aware models on different datasets. We see, first of all, that all multi-encoder models, including both Inside and Outside approaches outperform the sentence-level baselines by a large margin on the Zh-En and En-De datasets with a small $p$ value of $dropout$. Also, there are modest BLEU improvements on the Fr-En and En-Ru tasks. When the models are regularized by a larger $dropout$, all systems obtain substantial improvements - but the gaps between sentence-level and multi-encoder systems decrease significantly.

We deduce that if the context-aware systems rely on the contextual information from the preceding sentence, the performance of \textit{Random} and \textit{Fixed} should dramatically decrease due to the incorrect context.
Surprisingly, both \textit{Random} and \textit{Fixed} systems achieve comparable performance or even higher BLEU scores than \textit{Context} in most cases (See Table \ref{tab:main-results}). A possible explanation is that the context encoder does not only model the context. Instead, it acts more like a noise generator to provide additional supervised signals to train the sentence-level model.

\begin{table}[t!]
  \begin{center}
  \small
  \begin{tabular}{lcccc}
  \toprule
  \multicolumn{1}{c}{\multirow{2}{*}{\textbf{System}}}
  & \multicolumn{2}{c}{\textbf{Inside}}
  & \multicolumn{2}{c}{\textbf{Outside}} \\
  \cmidrule(lr){2-3} \cmidrule(lr){4-5}
  & Aware &  Agnostic & Aware &  Agnostic\\
  \midrule
  \textit{Context}    & 31.0 & 31.0  & 30.7  & 31.1 \\
  \textit{Random}     & 30.7 & 30.8  & 31.1  & 31.3 \\
  \textit{Fixed}      & 30.8 & 30.8  & 31.1  & 31.1 \\
  \bottomrule
  \end{tabular}
  \end{center}
\caption{The BLEU scores $[\%]$ of context-aware systems with two inference schemas. Aware represents the inference process matches the training. Agnostic represents that models ignore context encoder during inference.}\label{tab:inference}
\end{table}

\subsection{Robust Training}

To verify the assumption of robust training, we followed the work \cite{SrivastavaHKSS14,berger-etal-1996-maximum}. We turned off the context-encoder during the inference process, and made the inference system perform as the sentence-level baseline.
Table \ref{tab:inference} shows that both \textit{Context} and \textit{Random} inference without context-encoder obtain modest BLEU improvements.  This confirms that the information extracted by context-encoder just plays a role like introducing randomness into training (e.g., $dropout$), which is a popular method used in robust statistics. We argue that three types of context provide noise signals to disturb the distribution of the sentence-level encoder output. The BLEU improvements of both Outside and Inside are mainly due to the richer noise signals which can effectively alleviate the overfitting.

Inspired by Outside integration manner, we designed a simple yet effective method to regularize the training process: A Gaussian noise is added to the encoder output instead of the embedding \cite{cheng-etal-2018-towards}. We sample a vector $\epsilon \sim \mathbf{N}\left(0, \sigma^{2} \mathbf{I}\right)$ from a Gaussian distribution with variance $\sigma^2$, where $\sigma$ is a hyper-parameter. As seen in Table \ref{tab:gaussian}, the systems with Gaussian-noise significantly outperform the sentence-level baselines, and are slightly better than the Outside-context counterpart. Moreover, a natural question is whether further improvement can be achieved by combining the \textit{Context} with the Gaussian-noise method. From the last line in Table \ref{tab:gaussian}, we observe no more improvement at all. The observation here convinced the assumption again that the context-encoder plays a similar role with the noise generator.

\begin{table}[t!]
  \begin{center}
  \small
  \begin{tabular}{lcccc}
  \toprule
  \textbf{System} & \textbf{Zh-En} & \textbf{Fr-En} & \textbf{En-De} & \textbf{En-Ru} \\
  \midrule
  Baseline   & 19.7 & 36.9 & 30.2 & 31.1\\
  \textit{Context}    & 19.8 & 37.4 & 30.7 & 31.1 \\
  Noise   & 19.9 & 37.4 & 30.9 & 31.3\\
  \textit{Context}+Noise &19.9 & 37.3 & 30.9 & 31.3\\
  \bottomrule
  \end{tabular}
  \end{center}
\caption{Comparison of Outside \textit{Context} and Gaussian-noise methods on three tasks, with $dropout=0.3$, $\sigma$ = $0.3$.}\label{tab:gaussian}
\end{table}

 \subsection{Large Scale Training}

\begin{figure}[t!]
  \centering
  \begin{tikzpicture}
    \begin{axis}[
      width=7.0cm,height=4.0cm,
      xlabel=Data Volume,
      ylabel=BLEU,
      symbolic x coords={500k, 1M, 2M, 5M},
      ybar,enlargelimits=0.15,
      bar width=7.5pt,
      xlabel near ticks,
      ylabel near ticks,
      ylabel style={font=\footnotesize},
      xlabel style={font=\footnotesize},
      xticklabel style={font=\footnotesize},
      yticklabel style={font=\footnotesize},
      area legend,legend pos=north west,
        legend style={cells={anchor=west}, legend columns=-1},
      ]
    \pgfplotsset{
      legend image code/.code={
        \draw [#1] (0cm,-0.06cm) rectangle (0.4cm,0.1cm);
      },
     }
    \addplot[fill=mygreen!90, draw=mygreen!90]
    coordinates
    {
    (500k, 17.41) (1M, 20.61)
    (2M, 22.87) (5M, 24.1)
    };
    \addlegendentry{\scriptsize{Base}}

    \addplot[fill=myblue!90, draw=myblue!90]
    coordinates
    {
    (500k, 17.88) (1M, 20.81)
    (2M, 23.04) (5M, 24.51)
    };
    \addlegendentry{\scriptsize{Inside}}

    \addplot[fill=myred!90, draw=myred!90]
    coordinates
    {
    (500k, 18.11) (1M, 21.28)
    (2M, 23) (5M, 24.71)
    };
    \addlegendentry{\scriptsize{Gaussian}}

    \end{axis}
    \end{tikzpicture}
  \caption{BLEU scores vs. different data volume on Zh-En sentence-level dataset. $dropout=0.1$ and $\sigma=0.3$.}\label{fig:zh-en}
\end{figure}
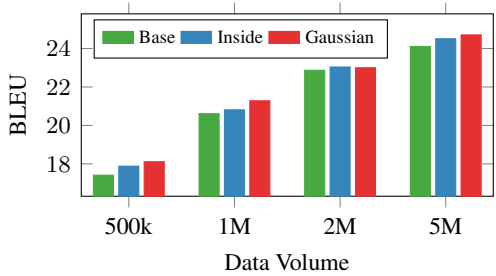

Most previous results are reported on small training datasets. Here we examine the effects of the noise-based method on different sized datasets. We trained the Inside-Random model and the Gaussian-noise model on different datasets consisting of 500K to 5M sentence pairs.  Seen from Figure \ref{fig:zh-en}, the baseline model achieves better translation performance when we increase the data size. More interestingly, it is observed that Inside-\textit{Random} and Gaussian-noise perform slightly better than the baseline, and the gaps gradually decrease with the volume increasing. This is reasonable that models trained on large-scale data may suffer less from
the overfitting problem.


\section{Related Work}

Context-aware NMT systems incorporating the contextual information generate more consistent and coherent translations than sentence-level NMT systems. Most of the current context-aware NMT models can be classified into two main categories, single-encoder systems \cite{tiedemann-scherrer-2017-neural} and multi-encoder systems \cite{journals/corr/JeanLFC17,voita-etal-2018-context,zhang-etal-2018-improving}. \citet{voita-etal-2018-context} and \citet{zhang-etal-2018-improving} integrated an additional encoder to leverage the contextual information into Transformer-based NMT systems. \citet{miculicich-etal-2018-document} employed a hierarchical attention network to model the contextual information. \citet{maruf-haffari-2018-document} built a context-aware NMT system using a memory network, and \citet{maruf-etal-2019-selective} encoded the whole document with selective attention network. However, most of the work mentioned above utilized more complex modules to capture the contextual information, which can be approximately regarded as multi-encoder systems.

For a fair evaluation of context-aware NMT methods, we argue that one should build a strong enough sentence-level baseline with carefully regularized methods, especially on small datasets \cite{kim-etal-2019-document, sennrich-zhang-2019-revisiting}.  Beyond this, \citet{bawden-etal-2018-evaluating} and \citet{voita-etal-2019-good} acknowledged that BLEU score is insufficient to evaluate context-aware models, and they emphasized that multi-encoder architectures alone had a limited capacity to exploit discourse-level context. In this work, we take a further step to explore the main cause, showing that the context-encoder acts more like a noise generator, and the BLEU improvements mainly come from the robust training instead of the leverage of contextual information. Additionally, \citet{cheng-etal-2018-towards} added the Gaussian noise to word embedding to simulate lexical-level perturbations for more robust training. Differently, we added the Gaussian noise to the encoder output which plays a similar role with context-encoder, which provides additional training signals.

\section{Conclusions}

We have shown that, in multi-encoder context-aware NMT, the BLEU improvement is not attributed to the leverage of contextual information. Even though we feed the incorrect context into training, the NMT system can still obtain substantial BLEU improvements on several small datasets. Another observation is that the NMT models can even achieve better translation quality without the context encoder. This gives us an interesting finding that the context-encoder acts more like a noise generator, which provides rich supervised training signals for robust training. Motivated by this, we significantly improve the sentence-level systems with a Gaussian noise imposed on the encoder output. Experiments on large-scale training data demonstrate the effectiveness of this method.

\section*{Acknowledgments}
This work was supported in part by the National Science Foundation of China (Nos. 61876035 and 61732005), the National Key R\&D Program of China (No. 2019QY1801) and the Opening Project of Beijing Key Laboratory of Internet Culture and Digital Dissemination Research. The authors would like to thank anonymous reviewers for their comments.

\bibliography{acl2020}
\bibliographystyle{acl_natbib}

\end{document}